\def\year{2021}\relax
\documentclass[letterpaper]{article} 
\usepackage{aaai21}  
\usepackage{times}  
\usepackage{helvet} 
\usepackage{courier}  
\usepackage[hyphens]{url}  
\usepackage{graphicx} 
\usepackage{natbib}  
\usepackage{caption} 
\usepackage{amsthm,amsmath,amssymb}
\usepackage{subfigure}
\usepackage{mathrsfs}
\title{An Unsupervised Sampling Approach for Image-Sentence Matching Using Document-Level Structural Information}

\author{

   Zejun Li,\textsuperscript{\rm 1}
   Zhongyu Wei,\textsuperscript{\rm 1,4}\thanks{Corresponding author} Zhihao Fan,\textsuperscript{\rm 1}
   Haijun Shan,\textsuperscript{\rm 2}
   Xuanjing Huang\textsuperscript{\rm 3}
}
\affiliations{

    \textsuperscript{\rm 1}School of Data Science, Fudan University, China\\
    \textsuperscript{\rm 2}Zhejiang Lab, China\\
    \textsuperscript{\rm 3}School of Computer Science, Fudan University, China\\
    \textsuperscript{\rm 4}Research Institute of Intelligent and Complex Systems, Fudan University, China\\

    \{20210980139, zywei, 18210980005\}@fudan.edu.cn, workingshan@163.com, xjhuang@fudan.edu.cn

}

\begin{document}

\maketitle


\begin{abstract}
 In this paper, we focus on the problem of unsupervised image-sentence matching. Existing research explores to utilize document-level structural information to sample positive and negative instances for model training. Although the approach achieves positive results, it introduces a sampling bias and fails to distinguish instances with high semantic similarity. To alleviate the bias, we propose a new sampling strategy to select additional intra-document image-sentence pairs as positive or negative samples. Furthermore, to recognize the complex pattern in intra-document samples, we propose a Transformer based model to capture fine-grained features and implicitly construct a graph for each document, where concepts in a document are introduced to bridge the representation learning of images and sentences in the context of a document. Experimental results show the effectiveness of our approach to alleviate the bias and learn well-aligned multimodal representations.
\end{abstract}

\section{Introduction}

\begin{figure}[t]
\centering
\includegraphics[width=0.9\columnwidth]{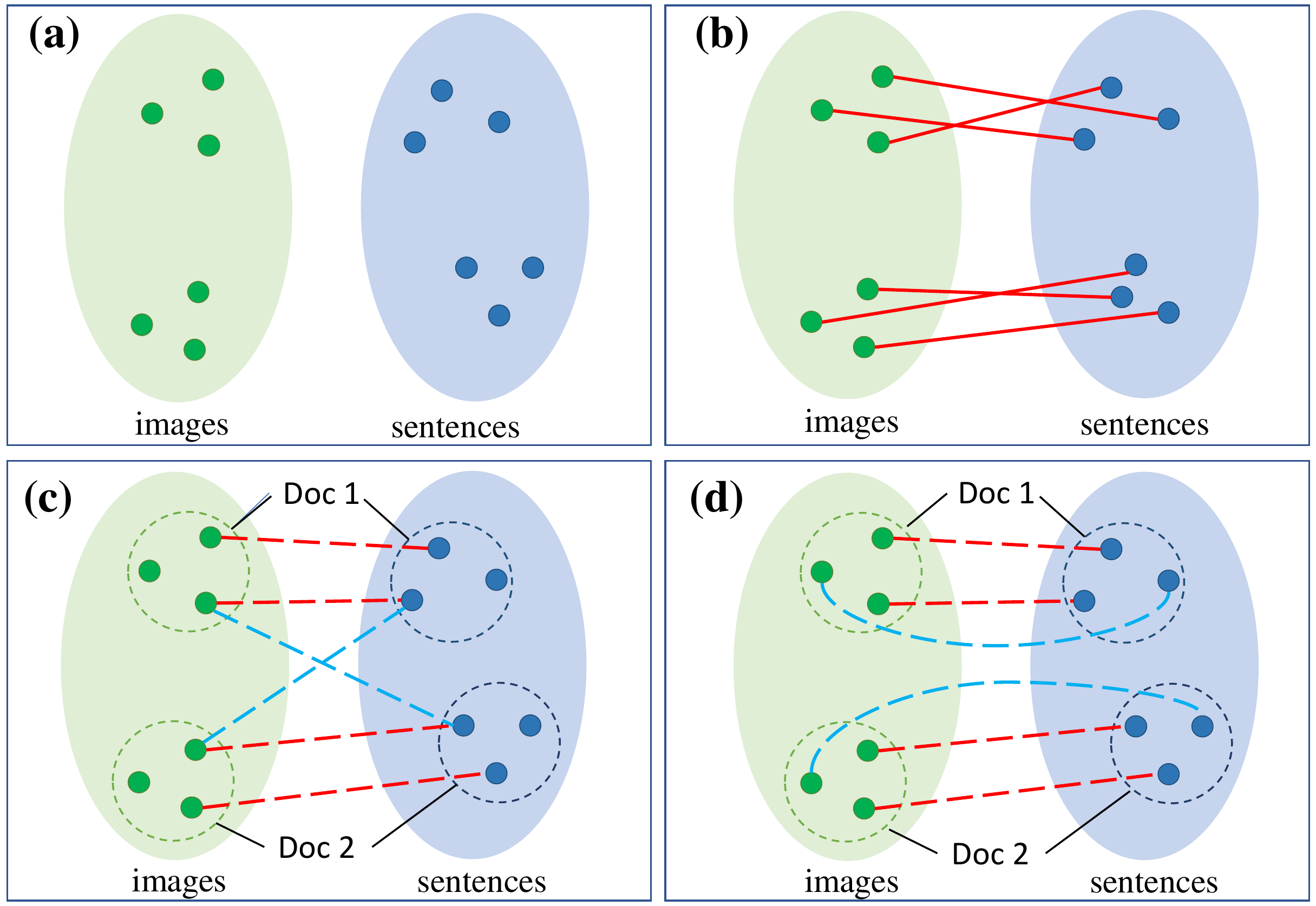} 
\caption{Illustration of different settings for image-sentence matching: (a) multimodal semantic spaces, (b) supervised alignment, (c) unsupervised cross-document objective in \cite{hessel-lee-mimno-2019unsupervised}, (d) our intra-document objective. Red links denote matched positive pairs, blue links denote negative pairs; solid links represent annotated labels, dashed lines represent pseudo labels detected by unsupervised methods; dashed circles denote image-sets and sentence-sets in documents.}
\label{fig1}
\end{figure}

Image-text matching is one of the fundamental problems in the field of vision and language~\cite{nam2017dual,huang2018learning}, and the main target is learning to align the semantic spaces of two modalities (Figure~\ref{fig1}(a)). Previous works on image-text matching is mainly supervised~ \cite{lee2018stacked,wang2019camp,zheng2020dual}, requiring large amounts of annotated image-sentence pairs (Figure~\ref{fig1}(b)). Considering labeled pairs of images and sentences are expensive to obtain, progress in developing unsupervised methods is therefore exciting and promising. Although some attempts are made to align image regions and segments of sentences~\cite{karpathy2014deep,karpathy2015deep,rohrbach2016grounding,datta2019align2ground}, they still rely on matched image-sentence pairs for distant supervision.

\begin{figure}[t]
\centering
\subfigure[Illustration of positive and negative samples for training and evaluation in \cite{hessel-lee-mimno-2019unsupervised}: links in red/green are negative/positive samples considered during evaluation, while links in yellow are negative samples considered during training.]{
\includegraphics[width=0.9\columnwidth]{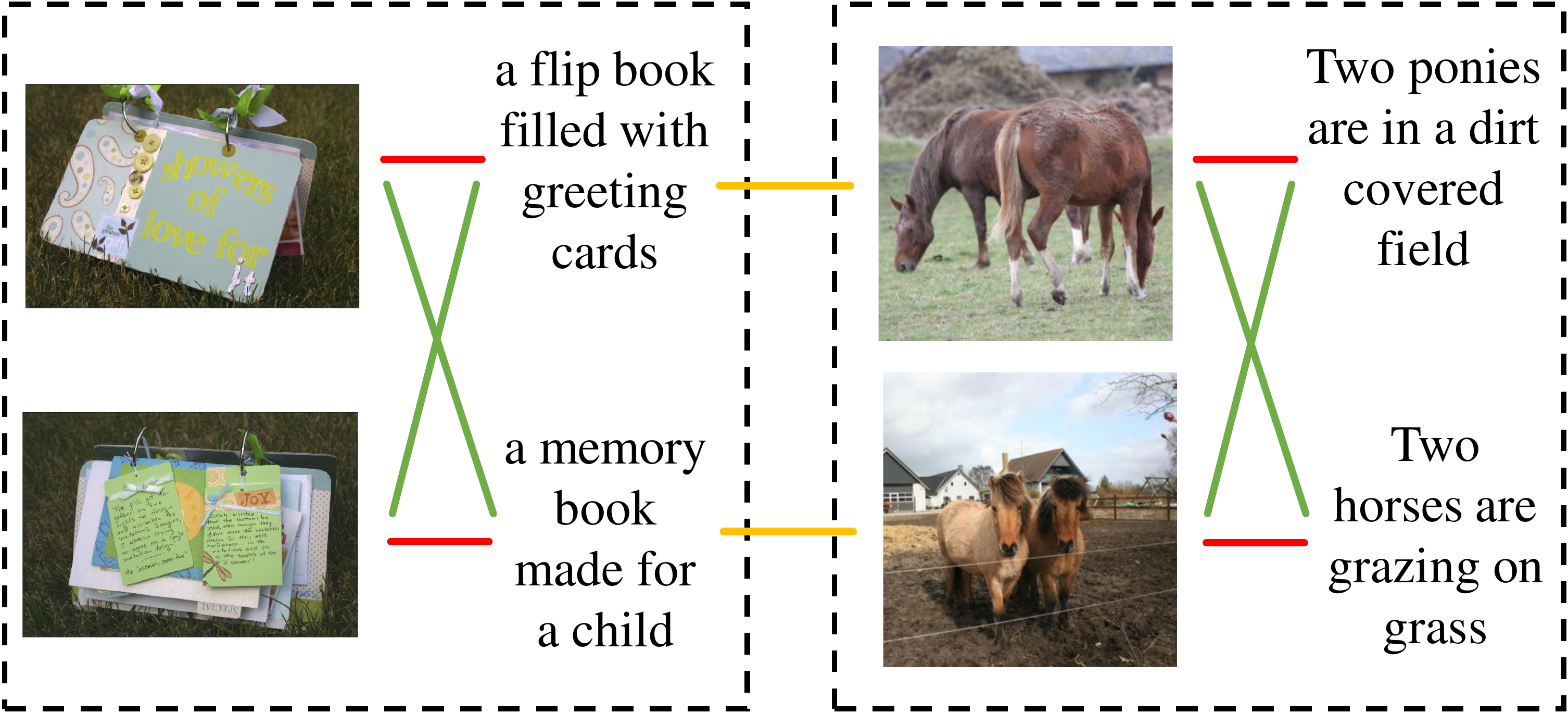} 
\label{bias_1}}
\subfigure[Distributions of L2 distances between pre-trained CNN features of ground-truth matched and negative images with respect to the same sentence, during inter-document training and intra-document evaluation.]{\includegraphics[width=0.9\columnwidth]{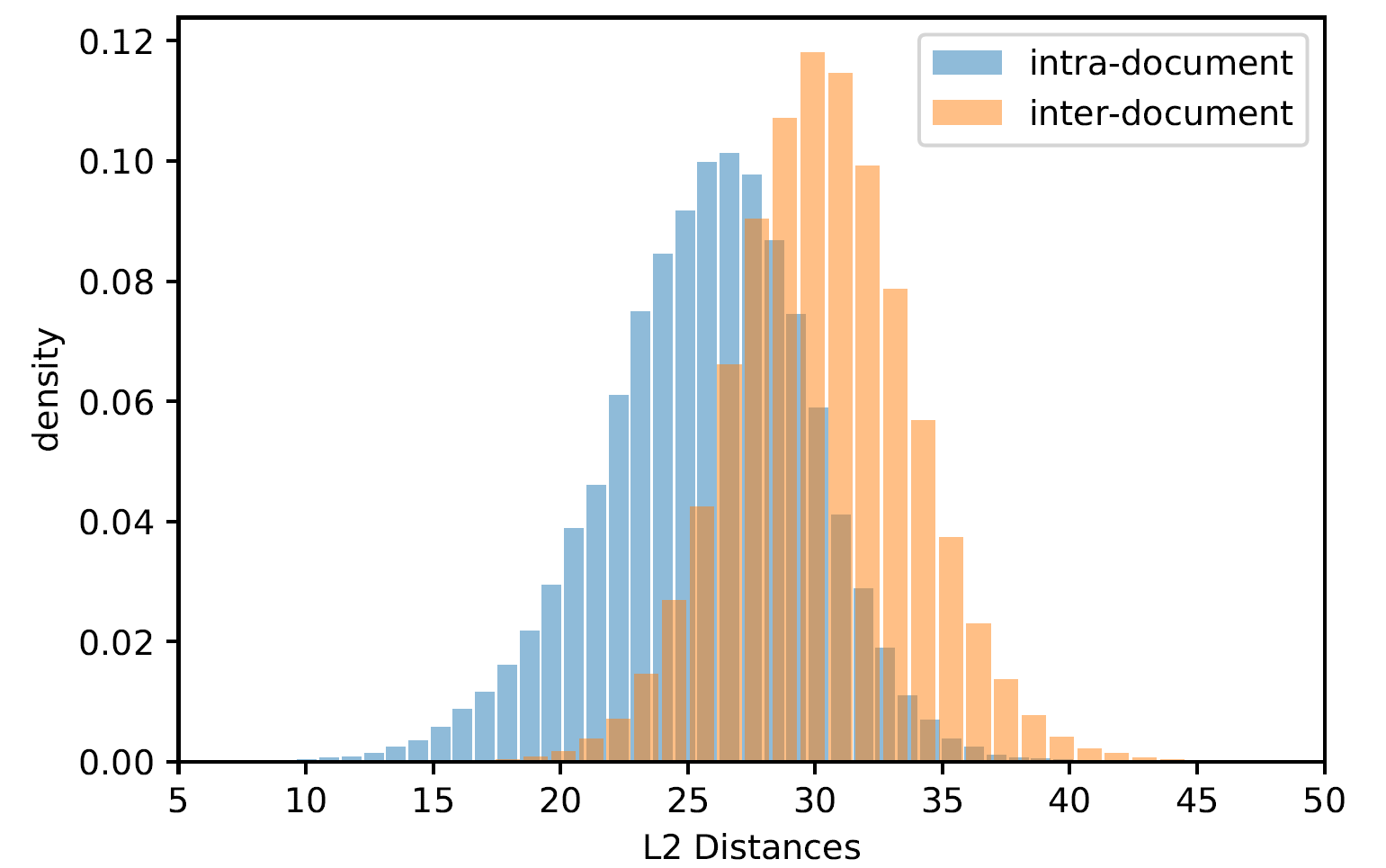} 
\label{bias}}
\caption{Illustration of the sampling bias.}
\end{figure}

The main challenge in unsupervised image-sentence matching is the lack of such information to distinguish positive and negative samples for model training. \cite{hessel-lee-mimno-2019unsupervised} explores to utilize the document-level structural information. In specific, image-sentence pairs in a document (intra-document) are selected as positive samples, and negative samples are drawn from cross-document image-sentence pairs (Figure~\ref{fig1}(c)). During training, the model is learning to maximize the distance between the sampled positive and negative pairs. Although this unsupervised sampling strategy is shown to be effective for learning aligned representations for images and sentences to some extent, the sampling bias between the training environment and real environment can not be ignored. 

Figure~\ref{bias_1} shows an example. The positive and negative sample pairs for training are much easier to be distinguished (book vs horses), while an image and a sentence in a negative sample from the testing environment can be highly correlated in terms of semantics. We further present the evidence for bias resulted in this sampling strategy in Figure~\ref{bias}. For each sentence in the VIST-DII dataset~\cite{hessel-lee-mimno-2019unsupervised}, we compute the L2 distances between pre-trained CNN features of its ground-truth matched and negative images considered in the training and evaluation process, then visualize the two distributions. The difference between these two distributions is significant\footnote{We conduct a two-sample Kolmogorov-Smirnov test where $p\text{-value}<0.01$.} and distances of training samples are generally larger. Such a sampling bias makes it hard for trained models to learn good representations for determining the correspondences between similar images and sentences.





To alleviate the issue of sampling bias, we propose to further distinguish positive and negative samples in the same document, which corresponds to Figure~\ref{fig1}(d). In practice, we sample negative samples from the pairs with the least semantic similarity in a document. Distinguishing similar images and sentences in a document requires the backbone model's capability to learn cross-modality representations with fine-grained information. Consider the document on the right in Figure~\ref{bias_1}, capturing the object-level details like ``dirt'' and ``grass'' is necessary to distinguish these 2 sentences and images. Motivated by the success of recent works on introducing concepts to bridge cross-modal learning~\cite{you2016image,fan2019bridging}, we further explore to model concepts in a document to bridge the semantics of images and sentences. We propose to extract concepts from images and build an intra-document graph composed of images, sentences, and concepts implicitly. A Transformer based model is utilized to model these implicit dependencies and represent images and sentences with context-encoded fine-grained features. 

The main contributions of our work are as follows:
\begin{itemize}

\item We reveal the sampling bias issue of an unsupervised sampling strategy for image-sentence matching that selects positive and negative samples from intra-document image-sentence and cross-document pairs, respectively.

\item To alleviate the sampling bias issue, we propose a strategy to select negative samples and additional positive samples from intra-document pairs and form a new objective for cross-modality representation learning. 
\item To recognize the complex pattern in intra-document samples, we propose a Transformer based model to capture fine-grained features and integrate concepts into our model to bridge the representation learning of multimodal data in a document.

\item We evaluate our method on the task of multi-model link prediction in multi-image, multi-sentence documents. Experiments show the effectiveness of our proposed method to alleviate the bias and learn better multimodal representations in the context of documents for this task. 
\end{itemize}


\section{Unsupervised Sampling Strategy based on Document-Level Structure}
\begin{figure}[t]
\centering
\includegraphics[width=0.9\columnwidth]{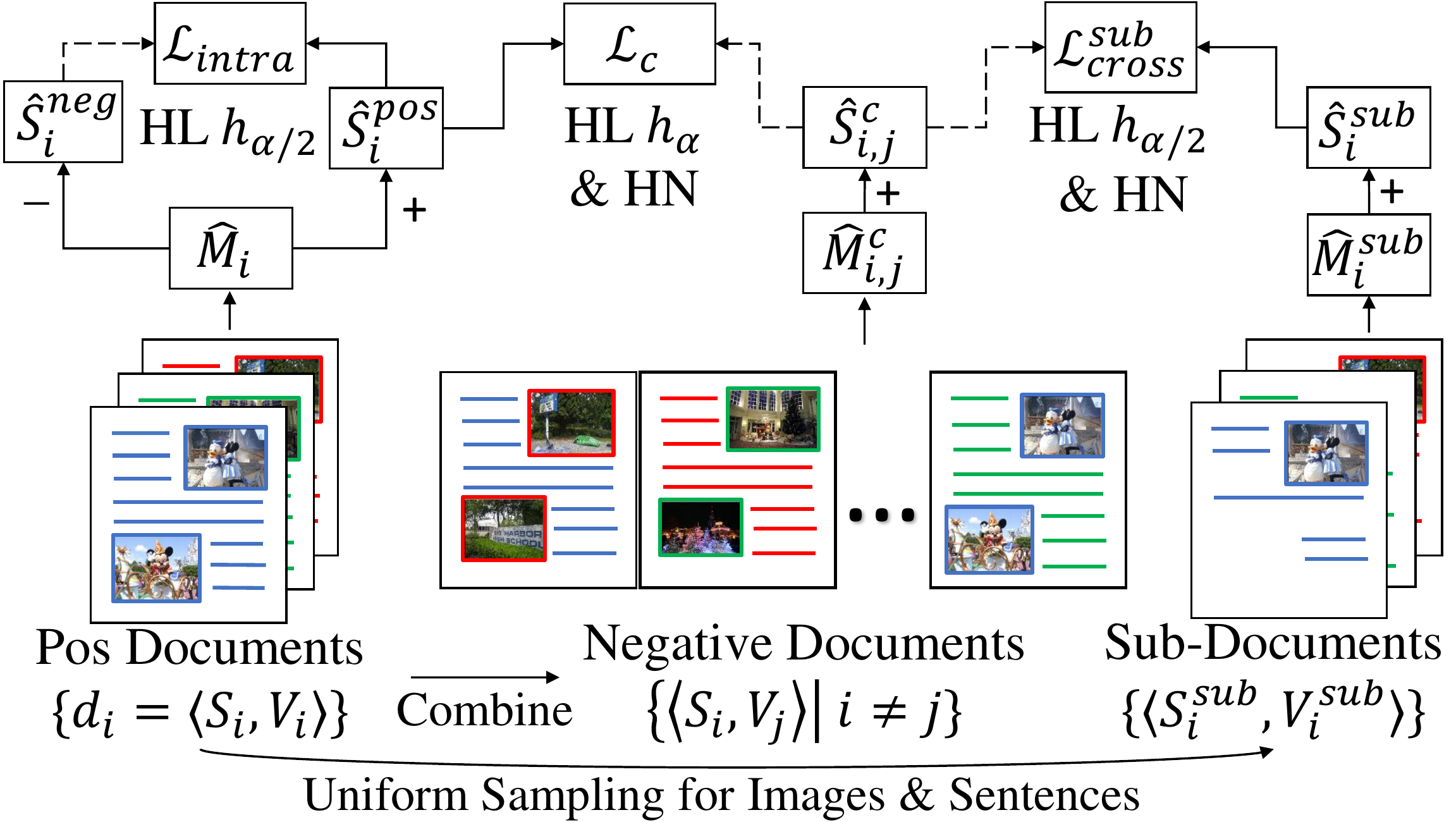} 
\caption{An illustration of the proposed training objectives, lines in documents represent sentences, different colors correspond to different documents. ``+'' represents the $\mathrm{TK}$ function and ``-'' represents the $\mathrm{NegTK}$ function. All $\hat{M}$ in the figure are computed by the same backbone model, and all $\hat{S}$ are the document-level similarities. Dashed lines indicate the negative inputs of hinge loss functions. ``HL'' and ``HN'' are short for hinge loss and hard negative, respectively.}
\label{fig2}
\end{figure}

We first introduce the setting of document-level structure proposed by~\citet{hessel-lee-mimno-2019unsupervised}. We are given a set of documents, each document $d_i=\langle S_i, V_i \rangle$ consists of a set $S_i$ of $|S_i|$ sentences and a set $V_i$ of $|V_i|$ images. On top of this, we define two kinds of image-sentence pairs, namely, intra-document pairs and cross-document pairs. Moreover, we propose three different strategies to sample positive or negative pairs and form three objectives namely, cross-document objective, intra-document objective, and dropout sub-document objective for model training. The overall illustration is shown in Figure~\ref{fig2}. $\hat{M}_{i}$ denotes the similarity matrix of intra-document pairs where each element $\hat{M}_{i,(m,n)}$ is the similarity between $S_{i,m}$ and $V_{i,n}$, $\hat{M}_{i,j}^c$ denotes the similarity matrix of cross-document pairs where each element $\hat{M}_{i,j,(m,n)}^c$ is the similarity between $S_{i,m}$ and $V_{j,n}$.

\subsection{Cross-Document Objective} 
The first objective is based on the assumption that co-occurring image-set/sentence-set pairs should be more similar than non-co-occurring image-set/sentence-set pairs. 

We construct negative documents by combining non-co-occurring image-set/sentence-set pairs, and the objective for a single positive document can be characterized by hard negative mining with hinge loss:
\begin{align}
    \mathcal{L}_c(S_i, V_i)=\max_{j\neq i }h_{\alpha}(\mathrm{sim}(S_i, V_i),\mathrm{sim}(S_i,V_j)) \nonumber \\ 
    +\max_{j\neq i}h_{\alpha}(\mathrm{sim}(S_i, V_i),\mathrm{sim}(S_j,V_i))
    \label{equation1}
\end{align}
where we consider $i,j$ in a mini-batch. $h_{\alpha}(m,n)=\max(0, n-m+\alpha)$ is the hinge loss function with a margin of $\alpha$, $\mathrm{sim}$ is a similarity function to compute the similarity between an image-set/sentence-set pair by mapping the predicted association matrix $\hat{M}$ between them to a real number, this function will select representative image-sentence pairs according to a specific criterion, and calculate the average similarity of selected pairs as the document-level similarity. We use a function $\mathrm{TK}:\mathbb{R}^{|S|\times|V|}\mapsto \mathbb{R}$ here, where the k most likely sentence-to-image and image-to-sentence edges will be selected based on currently predicted similarity, then compute the average similarity of selected pairs as the document-level similarity. This procedure corresponds to the equation: $\mathrm{sim}(S_i,V_j)=\hat{S}^c_{i,j}=\mathrm{TK}(\hat{M}^c_{i,j})$.



\subsection{Intra-Document Objective} 
The second objective aims to select negative image-sentence pairs from a document, with the assumption that similarities between predicted non-corresponding image-sentence pairs should be lower than predicted matched image-sentence pairs from the same document.

Similar to $\mathrm{TK}$ which measures the ``positive'' similarity, we also introduce a function $\mathrm{NegTK}$ to measure the document-level ``negative'' similarity i.e. how similar are those predicted non-corresponding images and sentences, $\mathrm{NegTK}$ will first select the k most unlikely sentence-to-image and image-to-sentence edges based on current predicted similarity, then calculate the average similarity of selected pairs as the document-level ``negative'' similarity. Then we can characterize this intra-document objective by a hinge loss between the document-level ``positive'' similarity and ``negative'' similarity:
\begin{align}
\hat{S}^{pos}_i=&\mathrm{TK}(\hat{M}_i) \nonumber \\
    \hat{S}^{neg}_i=&\mathrm{NegTK}(\hat{M}_i) \nonumber \\ 
    \mathcal{L}_{intra}&(S_i,V_i)=h_{\frac{\alpha}{2}}(\hat{S}^{pos}_i, \hat{S}^{neg}_i) 
\label{equation2}
\end{align}
where $h(\cdot)$ has the same definition as in Equation~\ref{equation1} but with a smaller margin $\frac{\alpha}{2}$. NegTK can be efficiently implemented with an equivalent definition: $\mathrm{NegTK}(\hat{M}_i)=-\mathrm{TK}(-\hat{M}_i)$.

In essence, adding this complementary objective is equivalent to those intra-document image-sentence pairs with low predicted similarity as negative samples. Generally, it is nearly impossible for all image-sentence pairs to have a semantic association, we believe this strategy will have a high probability to choose those image-sentence pairs without an edge between them in ground-truth.
 
 \begin{figure*}[t]
\centering
\includegraphics[width=0.8\textwidth]{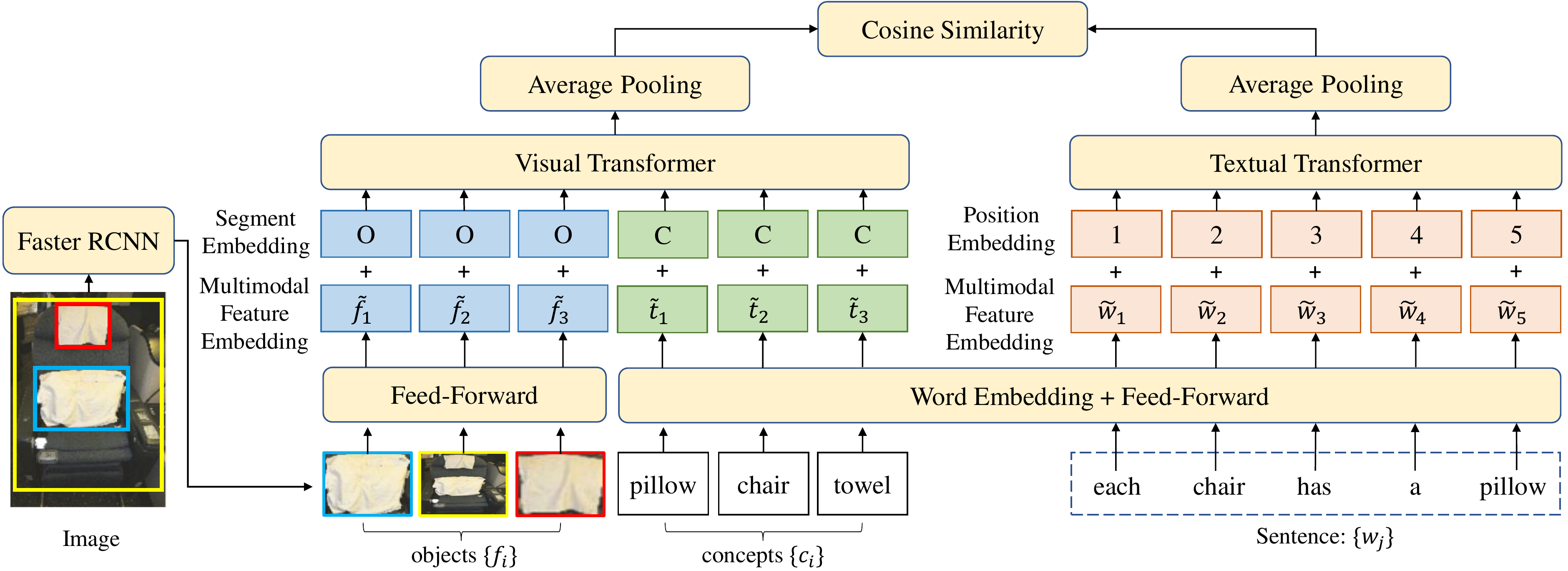} 
\caption{Architecture of our cross-modality alignment model.}
\label{model_fig}
\end{figure*}
 
\subsection{Dropout Sub-Document Objective} When using $\mathrm{TK}$, only the 2k most probable image-sentence pairs will be regarded as positive samples. Apart from those 2k selected edges, there may exist (weaker) semantic associations between other image-sentence pairs according to the composition of a document.

To utilize that information, we introduce another complementary cross-document objective, under the assumption that even if some sentences and pictures in a document are removed, the dropout sub-document composed of remaining sentences and images will also have a higher document-level similarity than those of negative documents, with a smaller gap. 

For a single positive document, we firstly construct a dropout sub-document by randomly removing a certain percentage ($1-p_{sub}$) of sentences and pictures:
\begin{align}
    S_i^{sub} = \mathrm{Uniform}(S_i,\mathrm{n}=\lfloor p_{sub}\times |S_i|\rfloor) \nonumber \\
    V_i^{sub} = \mathrm{Uniform}(V_i,\mathrm{n}=\lfloor p_{sub}\times |V_i|\rfloor)
    \label{equation3}
 \end{align}
where $\mathrm{Uniform}(A, \mathrm{n} = n_A)$ represents a function to uniformly draw $n_A$ samples from $A$ without replacement. Then we can characterize the objective with a form similar to $\mathcal{L}_c$:
\begin{align}
    \mathcal{L}_{cross}^{sub}(S_i, V_i)=\max_{i\neq j}h_{\frac{\alpha}{2}}(\mathrm{sim}(S_i^{sub}, V_i^{sub}),\mathrm{sim}(S_i,V_j)) \nonumber \\ 
    +\max_{i\neq j}h_{\frac{\alpha}{2}}(\mathrm{sim}(S_i^{sub}, V_i^{sub}),\mathrm{sim}(S_j,V_i))
\end{align}

At document level, adding this complementary objective means constructing more positive documents; at image-sentence pair level, it is equivalent to a new sampling strategy: select additional positive image-sentence pairs from a document without considering some images and sentences. But these positive samples are regarded weaker and supposed to have a smaller gap.

We combine 3 objectives to the total loss, for a single positive document, it will be:
\begin{align}
    \mathcal{L}(S_i,V_i)=\mathcal{L}_{c}(S_i,V_i)+\mathcal{L}_{intra}(S_i,V_i)+\mathcal{L}_{cross}^{sub}(S_i,V_i)
    \label{total_loss}
\end{align}

\section{Cross-Modality Alignment Model}

To better leverage information both in intra-document positive and negative image-sentence pairs, the backbone model needs to be able to represent images and sentences with fine-grained features and in the context of a document. Our model will extract representations of images and sentences in a $d_{\mathrm{multi}}$-dimensional multimodal text-image space. 

Our idea is to introduce another type of nodes ``concept'' into the original bipartite graph, these nodes represent possible concepts (entities) the document may contain, we utilize these nodes as an intermediary of implicit links between images and sentences, and bridge the representation learning of images and sentences in a document. Correct construction of such kind of graphs for every document is intractable, we thus resort to an implicit implementation with Transformer \cite{vaswani2017attention} -- which can be viewed as a densely connected graph model \cite{xu2019multi} -- and a shared embedding layer between words in sentences and concepts.

\subsection{Visual Objects and Concepts }
To extract possible concepts in a document, we consider visual entities detected from images since concepts in sentences are obscure and may have compound meanings. Following \citet{anderson2018bottom}, we use a pre-trained Faster RCNN \cite{ren2015faster} to extract $\mu$ object proposals $\{ o_1,...,o_{\mu}\}$ for each image, where each object $o_i$ is represented by its 2048-dimensional region-of-interest (ROI) feature $f_i$. At the same time, predicted labels of these objects are considered as extracted concepts $\{c_1,...,c_{\mu}\}$.

\subsection{Extracting Sentence Representations}
Similar to \citet{tan2019lxmert}, a sentence is first split into words $\{ w_1,...,w_{\lambda} \}$, then a word $w_j$ and its index $i$ ($w_i$'s absolute position in the sentence) are sent to a 300D word embedding layer and a position embedding layer respectively:
\begin{align}
    \Tilde{w}_i&=\mathrm{WordEmbed}(w_i) \nonumber \\
    \Tilde{u}_i&=\mathrm{PosEmbed}(i) \nonumber \\
    \Tilde{h}_i &= \mathrm{LayerNorm}(\Tilde{w}_i+\Tilde{u}_i)
    \label{wordemb}
\end{align}
where the word embedding layer is initialized with GoogleNews-pretrained word2vec embedding \cite{mikolov2013distributed}, the position embedding layer is randomly initialized from a uniform distribution between -0.02 and 0.02. 

We further project the embeddings with a feed-forward layer and encode them by a single-modality $N_T$-layer Transformer:
\begin{align}
    h_i^{0}&=W_T\Tilde{h}_i+b_T \nonumber \\
    \{h_1^{l+1},...,h^{l+1}_{\lambda}\}&=\mathrm{Transformer}^l_{T}( \{h_1^{l},...,h^l_{\lambda}\})
\end{align}
where $W_T\in\mathbb{R}^{d_{\mathrm{multi}}\times 300}$ and $b_T\in\mathbb{R}^{d_{\mathrm{multi}}}$ are parameters of the feed-forward layer.

Finally, we extract the representation of a sentence through an average pooling layer on top of $\{h_1^{N_T},...,h_{\lambda}^{N_T}\}$.

\subsection{Extracting Image Representations} A cross-modality Transformer is used to model the dependency between extracted visual objects and visual concepts in an image, modeling the links between concepts and images.

In this cross-modality Transformer, each object $o_j$ is represented by its ROI feature $f_j$ and a segment embedding indicating this token is a visual object:
\begin{align}
    \Tilde{f}_j&=\mathrm{LayerNorm}(W_Vf_j+b_V) \nonumber \\
    \Tilde{s}_j&=\mathrm{LayerNorm}(\mathrm{SegEmbed}(o_j)) \nonumber \\
    \Tilde{v}_i &= (\Tilde{f}_i+\Tilde{s}_i)/2
\end{align}

Each concept $c_k$ is sent to a word embedding layer and then projected through a feed-forward layer to learn a textual concept embedding, and a segment embedding is added to it to indicate this token is a visual concept:
\begin{align}
    \Tilde{w}_k&=\mathrm{WordEmbed}(c_k) \nonumber \\
    \Tilde{t}_k&=\mathrm{LayerNorm}(W_T\Tilde{w}_k+b_T) \nonumber \\
    \Tilde{s}_k&=\mathrm{LayerNorm}(\mathrm{SegEmbed}(c_k)) \nonumber \\
    \Tilde{c}_k &= (\Tilde{t}_k+\Tilde{s}_k)/2
\end{align}
If a concept consists of several words, an average pooling is applied to get the textual embedding. The word embedding layer shares weight with the layer in extracting sentence representations, as in Equation~\ref{wordemb}. This characterizes the links between concepts and sentences when concepts are directly mentioned in those sentences.

Then all objects and concepts in each image are sent into a $N_C$-layer cross-modality Transformer:
\begin{align}
    \{e_1^0,...,e_{2\mu}^0\}&=\{\Tilde{v_1},...,\Tilde{v}_{\mu},\Tilde{c}_1,...,\Tilde{c}_{\mu}\} \nonumber \\
    \{e_1^{l+1},...,e_{2\mu}^{l+1}\}&=\mathrm{Transformer}^l_{T}( \{e_1^l,...,e_{2\mu}^{l}\})
\end{align}
Similarly, an average pooling layer on all objects and concepts outputted by the last Transformer layer $\{e_1^{N_C},...,e_{2\mu}^{N_C}\}$ is used to extract the representation of an image.

\section{Experiments}

We evaluate our unsupervised training strategy on the task of multi-model link prediction in multi-image, multi-sentence documents proposed in \cite{hessel-lee-mimno-2019unsupervised}. 

\subsection{Multi-image Multi-sentence Linking}

For a document $d_i=\langle S_i, V_i \rangle$ consists of a set $S_i$ of $|S_i|$ sentences and a set $V_i$ of $|V_i|$ images, we aim to predict the label for each pair of image and sentence within the document. We generate features for all images and sentences based on trained cross-modality representation learning model and compute the similarity matrix $\hat{M}_i$ where the $(i,j)^{th}$ element is the cosine similarity between the $i^{th}$ sentence representation and $j^{th}$ image representation.

\subsection{Experiment Datasets}
We evaluate our proposed method on MSCOCO \cite{lin2014microsoft} and VIST \cite{huang2016visual}. \citet{hessel-lee-mimno-2019unsupervised} collect images and sentences to compose documents from those crowdlabeled dataset, then construct 3 different datasets for intra-document link prediction: MSCOCO, DII, and SIS. In MSCOCO, each document consists of 5 randomly sampled image-caption pairs, 5 distractor images, and 5 distractor sentences. In DII and SIS, each document consists of 5 images from the same album and 5 sentences of the corresponding description-in-isolation (DII) or story-in-sequence (SIS) story. Statistics of these datasets are given in Table 1.

\begin{table}
\begin{center}
\begin{tabular}{lcccc}
\hline
           & \multicolumn{1}{l}{train/val/test} & \multicolumn{1}{l}{$n_i/m_i$} & \multicolumn{1}{l}{\# imgs} & \multicolumn{1}{l}{density} \\ \hline
MSCOCO     & 25K/2K/2K                       & 10/10                         & 83K                      & 5\%                      \\
Story-DII  & 22K/3K/3K                       & 5/5                           & 47K                      & 20\%                     \\
Story-SIS  & 37K/5K/5K                       & 5/5                           & 76K                      & 20\%                     \\
\hline
\end{tabular}
\end{center}
\caption{Dataset Statistics: density refers to edge density in the ground-truth graph for a document, i.e. the number of ground-truth edges divided by the number of all possible edges ($n_i*m_i$)}
\label{table1}
\end{table}

\subsection{Implementation Details}

For images, we use a Faster-RCNN pre-trained on Visual Genome \cite{krishna2017visual} provided by \cite{anderson2018bottom}. In DII and SIS, we extract 36 objects and concepts for each image, while the number is adaptive in MSCOCO. We set $d_{multi}=1024$, $N_T=3$ and $N_C=3$. Each layer in the single-modality Transformer and cross-modality Transformer has 8 heads. Mask in Transformers is used to deal with sequences with variable lengths. For training objective, in function TK, we set $k=\mathrm{min}(n_i,m_i)$. The sub-document proportion $p_{sub}$ is set as 0.6 in SIS and DII, 0.8 in MSCOCO. Margin $\alpha$ in hinge loss is set to 0.2. We train our model using Adam optimizer \cite{kingma2014adam}. With a warm-up phase, we linearly increase the learning rate from 1e-7 to the configured max learning rate after several steps. In MSCOCO, max learning rate and warm-up steps are set as 1e-5 and 3000 respectively, while they are set as 5e-5 and 4000 in SIS and DII. After the warm-up phase, we decrease the learning rate by a factor of 5 each time the total loss over the
validation set plateaus for more than 3 epochs. The mini-batch size is 11.

\subsection{Overall Performance}

\begin{table*}[t]
\centering

\begin{tabular}{l|cccccc}
\hline
            & \multicolumn{2}{c}{MSCOCO}                  & \multicolumn{2}{c}{Story-DII}               & \multicolumn{2}{c}{Story-SIS}               \\
            & AUC           & \multicolumn{1}{l}{p@1/p@5} & AUC           & \multicolumn{1}{l}{p@1/p@5} & AUC           & \multicolumn{1}{l}{p@1/p@5} \\ \hline
Obj Detect  & 89.5          & 67.7/45.9                   & 65.3          & 50.2/35.2                   & 58.4          & 40.8/28.6                   \\
NoStruct    & 87.4          & 50.6/34.3                   & 77.0          & 60.8/46.3                   & 64.5          & 42.8/33.2                   \\ \hline
MulLink     & 99.0 & 95.0/81.1          & 82.9          & 72.0/55.8                   & 68.8          & 51.8/38.6                   \\
Ours & \textbf{99.3} & \textbf{97.6}/\textbf{86.0}                   & \textbf{85.5} & \textbf{77.2}/\textbf{60.1}          & \textbf{70.2} & \textbf{53.1/39.8}                    \\ \hline
\end{tabular}

\caption{Overall performance of different models. Numbers in bold denote the best performance in each column.}
\label{table2}
\end{table*}

We compare our model with the model proposed in \cite{hessel-lee-mimno-2019unsupervised}, which are the only existing unsupervised model for this task, 2 baseline models are also proposed by \citet{hessel-lee-mimno-2019unsupervised} and listed here for a comparison:
\begin{itemize}
    \item \textbf{Object Detection} Each image is represented by the average of the word2vec embeddings of its $K$ most probable labels predicted by pretrained DenseNet169 \cite{huang2017densely}, while each sentence is represented by the average of the word2vec embeddings of its words.
    \item \textbf{NoStruct} randomly samples image-caption pairs from a document and treat the similarity between them as the document-level similarity.
    \item \textbf{MulLink} \cite{hessel-lee-mimno-2019unsupervised} uses a GRU-CNN based backbone model to encode images and sentences, and it is trained only using the loss in Equation~\ref{equation1}. 
\end{itemize}

Table~\ref{table2} shows the comparative results. Both MulLink and our model show a superior performance than baseline models, reflecting their capability to measure cross-modality similarity more efficiently and sample effective image-sentences pairs in a structural document under the unsupervised setting. In general, Our approach outperforms MulLink in all 3 datasets, which means our proposed sampling strategy and fine-grained backbone model help to improve the performance jointly. 

In MSCOCO, there is nearly no bias between intra-document and cross-document negative image-sentence pairs, due to the composition of documents, so MulLink has already achieved nearly perfect performance on the AUC metric. Our model still shows superior performance on $P@1$ and $P@5$, which means MulLink gets troubled in distinguishing some negative intra-document pairs, and our backbone model has a stronger ability to measure cross-modality in finer granularity. In DII, images and sentences in a document are more similar, i.e. a larger differences between cross-document and intra-document image-sentence pairs. In such a setting, distinguishing positive and negative intra-document samples requires fine-grained cross-modality similarity measurement, our approach therefore has obvious improvements on all metrics. In SIS, the main challenge is how to understand each sentence in a story, sentences are tightly related to each other and may even use pronouns to refer to words in other sentences. Our model still outperforms MulLink.

\begin{table}[]
\centering
\begin{tabular}{l|c|cc}
\hline
\multicolumn{1}{c|}{backbone} & Objectives        & AUC           & p@1/p@5            \\ \hline
\textbf{1 Ours}          & \textbf{C+I+D} & \textbf{85.5} & \textbf{77.2/60.1} \\ \hline
2 w/o Concept       & C+I+D          & 85.3          & 75.8/59.8          \\
3 w/o T               & C+I+D          & 85.1          & 75.0/59.0          \\
4 w/o T\&Concept      & C+I+D          & 85.1          & 74.6/59.1          \\
5 GRU+CNN                       & C+I+D         & 84.0          & 72.9/58.0            \\ \hline
6 Ours                   & C+I          & 85.2          & 75.9/59.2          \\
7 Ours       & C+D          & 85.4          & 76.2/59.9          \\
8 Ours               & I+D          & 84.1          & 73.4/57.8          \\
9 Ours            & C              & 85.0           & 75.5/59.4 \\\hline
\end{tabular}
\caption{Ablation study on SIS, the ``Objectives'' column represents different combinations of objectives used during training, where ``C'', ``I'', and ``D'' correspond to 3 parts of objectives mentioned, respectively. ``T'' is short for Transformer, $w/o$ means removing a certain module.}
\label{table3}
\end{table}

\begin{figure}[t]
\centering
\includegraphics[width=0.9\columnwidth]{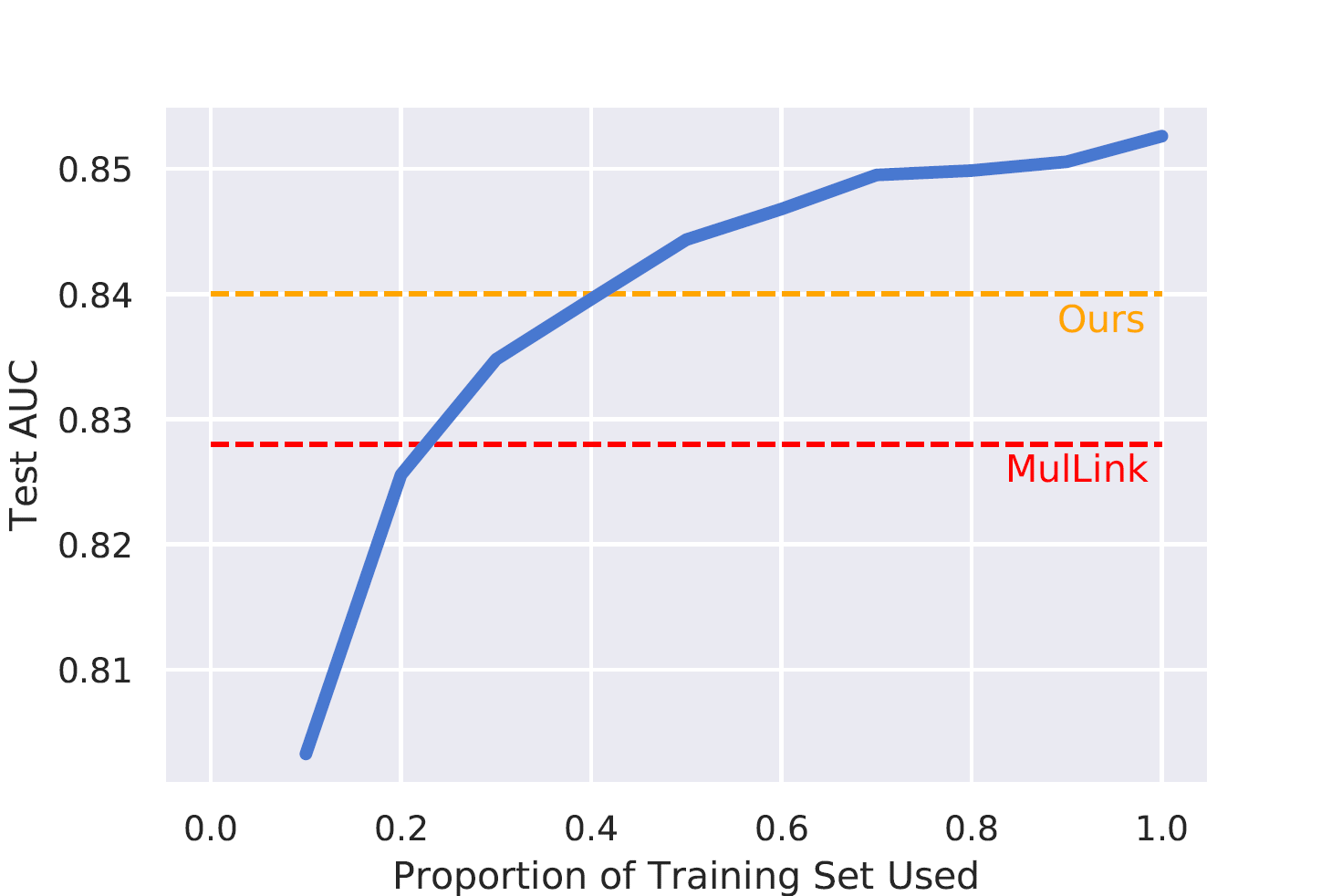} 
\caption{Performance of supervised strategy using different proportions of training data, dashed lines denote performances of unsupervised strategies.}
\label{efficiency}
\end{figure}

\begin{figure*}[t]
\centering
\includegraphics[width=0.95\textwidth]{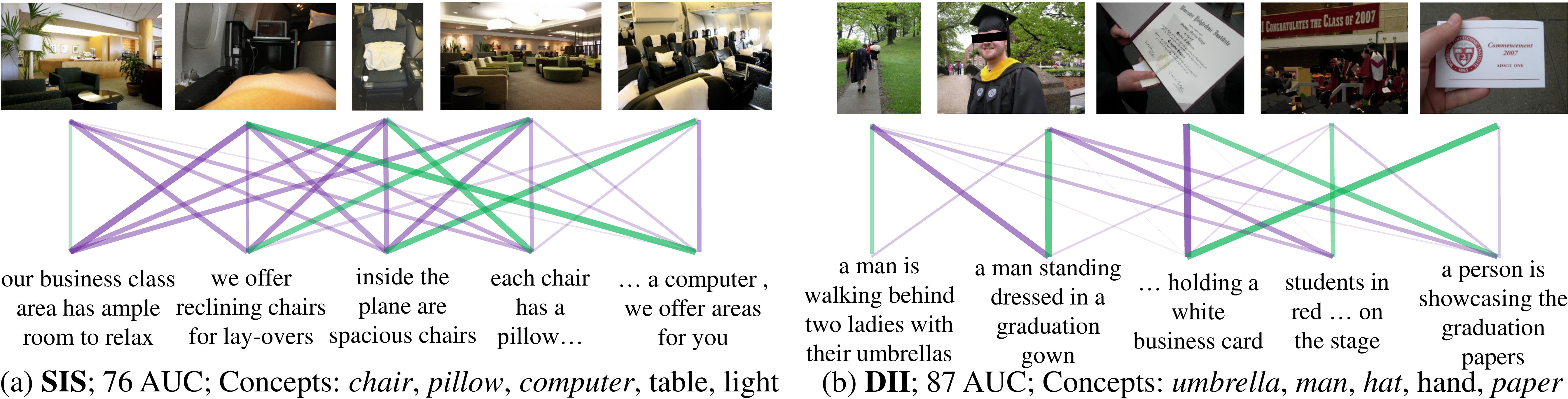} 
\caption{Illustrative documents in DII and SIS: Edges in green are true links in ground-truth; edge widths show the magnitude of edges in $\hat{M}_i$ (only positive weights are shown). Main detected concepts are listed and \emph{italicized words} are directly involved in sentences. Selected documents are representative because their AUC scores match the average AUC in corresponding datasets.}
\label{fig3}
\end{figure*}

\subsection{Ablation Study} To better understand the influence of each module in our approach, we perform the ablation study on the DII dataset, and results are shown in Table~\ref{table3}. In those variations without Transformer, we use a softmax pooling to aggregate all objects (and concepts) features to represent an image, where weights for softmax are computed by a linear layer, w/o concepts means only a sequence of object features are sent to backbone models. Several findings stand out:
\begin{itemize}
    \item Generally, it is showed that each objective contributes to the performance. Cross-document objective (``C'') is the main part since it directly leverages document-level co-occurrence information, other 2 are supplementary objectives to sample more examples with respect to reasonable assumptions, therefore the performance is not satisfying when only using 2 supplementary objectives (see row 8). Intra-document objective (``I'') helps to alleviate the bias, dropout sub-document objective (``D'') aims to introduce randomness and discover weak cross-modal association, both of them utilize more information and enhance the performance, and the combination of 3 objectives helps the model reach the best performance.
    \item Without Transformer, just aggregating the concept features into the image representation does not improve performance (see row 2, 3), showing that the implicit graph between concepts and objects modeled by Transformer is necessary to extract better image representations. 
    \item Incorporating concepts into Transformer significantly improves performance on precision (see row 1, 2). Illustrating that modeling of the dependency between objects and concepts is effective. An intuitive case is that our model will easily detect the cross-modal association if sentences involve classes of objects that appear in images.
\end{itemize}

\section{Further Analysis}
\subsection{Bias Alleviation} Our proposed sampling strategy aims to alleviate the bias between cross-document training and intra-document evaluation, we conduct error analysis to show the effectiveness of our approach more intuitively. As the "spread" hypothesis in \cite{hessel-lee-mimno-2019unsupervised}, documents with lower diversity among images/sentences are harder to disambiguate at test time. This hypothesis corresponds to our idea, lower intra-document diversity is equivalent to larger bias between intra-document and cross-document image-sentence pairs, since cross-document image-sentence pairs are always totally uncorrelated. 

So we follow the error analysis setting for the "spread" hypothesis, we use DenseNet169 features for images and mean word2vec for sentences, then compute the mean squared distance to their centroid to quantify the spread of a document. An OLS regression of image
spread + text spread on test AUC scores is fitted and its R-Square statistic shows how much of the variance in AUC can be explained by the intra-document spread. For DII and SIS, our approach reduces the R-Square from 42\% to 26\% and 23\% to 12\% respectively. This experiment does not involve MSCOCO since AUC scores are all large. 

These results illustrate that our approach weakens the influence of intra-document diversity (bias between training and evaluating). Accompanied by the superior overall performance, it is strong proof of our approach's effectiveness to alleviate the bias, under the unsupervised setting.

\begin{table}[]
\centering
\begin{tabular}{l|cc}
\hline
\multicolumn{1}{c|}{Method} & AUC  & p@1/p@5   \\ \hline
1 Transfer from MSCOCO        & 78.6 & 66.5/49.5 \\
2 Unsupervised                & 85.5 & 77.2/60.1 \\
\hline
\end{tabular}
\caption{Performance of different methods on DII without explicit labels.}
\label{table4}
\end{table}



\subsection{Comparison with Supervised Strategy}

To show the efficiency of our unsupervised sampling strategy, we compare the performance with a supervised model, CNN-RNN is used as the backbone model. We vary the proportion of samples used to train supervised models and present the results in Figure~\ref{efficiency}. It reveals that the difference between fully trained supervised and unsupervised strategies is not that large. And it needs more than 40\% of samples for the supervised strategy to generate better performance than our unsupervised approach (20\% to beat \emph{MulLink}). 


In addition, we compare our model with a supervised model in the setting of transfer learning. We train the Transformer-based model on MSCOCO with ground-truth image-sentence pairs and test it on DII. Results can be seen in Table~\ref{table4}, without ground-truth labels in the target domain, our unsupervised method shows a better performance.


\subsection{Case Study} To show the effectiveness of using more information provided by additional intra-document samples and appropriate model architecture, we present two illustrative examples in Figure~\ref{fig3}, the form of illustration is the same as in \cite{hessel-lee-mimno-2019unsupervised}. It shows that our models are able to discover fine-grained association by detecting and utilizing objects and corresponding concepts.


\section{Related Work}

Image-sentence matching is one of the fundamental tasks in the field of vision and language \cite{nam2017dual,huang2018learning}. A rich line of early studies focus on one-to-one matching \cite{yan2015deep,klein2015associating,faghri2017vse++,gu2018look}, usually extract global representations for image and sentence, then measure their similarities in a joint semantic space through. With the success of deep learning, employing CNN and RNN as modality-specific encoders becomes the mainstream. To learn an aligned multimodal semantic space where matched image-sentence pairs have small distances or high similarities, proposed training strategies usually use triplet ranking loss \cite{yan2015deep,kiros2014unifying,klein2015associating,peng2019cm}, while hard negative mining is showed to significantly improve the performance in \cite{faghri2017vse++}.

To capture fine-grained cross-modality association, most existing many-to-many matching methods try to incorporate relationships between image regions and sentence words \cite{karpathy2014deep, karpathy2015deep,huang2017instance,lee2018stacked,wu2019learning}. Some works align image segments and portions of a sentence without explicit labels \cite{karpathy2014deep,karpathy2015deep, rohrbach2016grounding, datta2019align2ground}.

Generally, most previous works follow a retrieval paradigm within a large dataset \cite{lin2014microsoft,young2014image}, where images and sentences are independent. \citet{hessel-lee-mimno-2019unsupervised} formulate the task of multimodal intra-document links prediction in multi-image multi-sentence documents, some documents are collected from the datasets of visual storytelling, which is another task requiring modeling for intra-document dependency~\cite{huang2016visual,wang2020storytelling}.

\section{Conclusion and Future Work}

In this work, we focus on the problem of unsupervised image-sentence matching. In order to alleviate the sampling bias introduced by the existing unsupervised training strategy, we propose a new sampling strategy to efficiently sample additional positive and negative intra-document samples. In addition, we propose to use a Transformer based model to learn cross-modality representations for images and sentences. Our approach improves the matching accuracy of an unsupervised multimodal link prediction task across different datasets. In the future, we would like to explore more downstream tasks using our unsupervised matching strategy. Besides, it is interesting to investigate few-shot semantic concept detection in an unsupervised way.

\section{Acknowledgments}

This work is partially supported by Ministry of Science and Technology of China (No.2020AAA0106701), Science and Technology Commission of Shanghai Municipality Grant (No.20dz1200600, 17JC1420200). We would also like to thank Ruize Wang and the anonymous reviewers for their constructive feedback.

\bibliography{paper.bib}

\end{document}